\documentclass[conference]{IEEEtran}
\newtheorem{definition}{Definition}
\IEEEoverridecommandlockouts

\usepackage{cite}
\usepackage{amsmath,amssymb,amsfonts}
\usepackage{algorithmic}
\usepackage{graphicx}
\usepackage{textcomp}
\usepackage{xcolor}
\usepackage{makecell}
\usepackage{hyperref}
\usepackage[linesnumbered,lined,ruled]{algorithm2e}
\def\BibTeX{{\rm B\kern-.05em{\sc i\kern-.025em b}\kern-.08em
    T\kern-.1667em\lower.7ex\hbox{E}\kern-.125emX}}

\begin{document}
\title{CLOE: Christoffel Loss Autoencoder for Anomaly Detection}

\author{
\IEEEauthorblockN{
Léa Billet\IEEEauthorrefmark{1}\IEEEauthorrefmark{2},
Louise Travé-Massuyès\IEEEauthorrefmark{1},
Elodie Chanthery\IEEEauthorrefmark{1},
Alexandre Gaffet\IEEEauthorrefmark{2}
}

\IEEEauthorblockA{
\IEEEauthorrefmark{1}
LAAS-CNRS, Université de Toulouse, CNRS, INSA, Toulouse, France\\
\{lea.billet, louise, elodie.chanthery\}@laas.fr
}

\IEEEauthorblockA{
\IEEEauthorrefmark{2}
SCHAEFFLER, Toulouse, France\\
\{lea.billet, alexandre.gaffet\}@mail.schaeffler.com
}
}

\maketitle
\begin{abstract}
    Semi-supervised anomaly detection plays a key role in diverse fields such as process monitoring, healthcare, and finance. However, lightweight methods often struggle with high-dimensional data and typically require careful tuning of multiple hyperparameters. Among existing approaches, Christoffel Function–based methods are attractive due to their simplicity, requiring at most a single hyperparameter. They also benefit from a well-established theoretical foundation that yields several interesting results for data science. However, their main limitation is poor scalability to high-dimensional settings. In this paper, we introduce CLOE, a new method that combines an autoencoder for dimensionality reduction with a Christoffel Function–based detector applied in the latent space. To better align representation learning with anomaly detection, we design a novel loss function that leverages the Christoffel Function to guide the autoencoder toward representations that better capture the support of the normal data distribution. We further propose a principled procedure to set the detection threshold and an efficient strategy to tune the single remaining hyperparameter. Experiments on multiple high-dimensional tabular anomaly detection benchmarks demonstrate that CLOE achieves superior performance compared to existing methods, while preserving the lightweight and low-tuning advantages of Christoffel Function–based approaches.
\end{abstract}


\section{Introduction}
The growth in sensor deployment for monitoring activities in health, industry, and other domains is creating substantial amounts of high-dimensional data. A crucial application is anomaly detection (AD), i.e., identifying abnormal or rare events, known as outliers. In semi-supervised learning, AD methods are trained using samples known to be normal (inliers). These methods estimate the data distribution and compute a score for each test sample. To detect outliers, the score is compared to a threshold provided by the method~\cite{platt2001estimating}. However, most classical AD methods are challenged by the curse of dimensionality and do not consider the full complexity of data. The time complexity to estimate a distribution is very high, not always suitable for nonlinear settings, and cross-variable dependencies are not taken into account~\cite{samariya2023comprehensive} and~\cite{pang2021deep}. Among the various methods, those based on the Christoffel Function (CF) have drawn our attention~\cite{ducharlet2024leveraging}. Rooted in approximation theory and orthogonal polynomials, the CF is grounded in a rigorous algebraic framework that addresses key requirements of data science~\cite{lasserre2022christoffel}, particularly the need to be free from hyperparameter tuning~\cite{ducharlet2024leveraging}.

Deep learning offers a solution to handle high-dimensional data. A neural network can reduce the dimensionality of the data while considering cross-variable dependencies. Autoencoders (AE), a class of neural networks, consist of an encoder and a decoder that are trained to reconstruct the input data while reducing the data dimensionality in the latent space in a nonlinear way \cite{wang2016auto}. The encoder hence encodes data in a low-dimensional space so that a classical AD method can be used to detect outliers using the latent space. However, the learned representations may not optimally capture the support of the normal data for anomaly detection. To address this, the training of the autoencoder can be guided by the anomaly detection method, ensuring that the latent space provides more informative and discriminative representations. This principle is known as coupled or joint training \cite{huang2025deep}.

In this paper, we propose CLOE (Christoffel LOss for autoEncoder), an efficient approach for high-dimensional tabular anomaly detection in a one-class classification setting, i.e., only normal samples are available during training. In CLOE, an AE reduces data dimensionality, and its latent space is regularized using the empirical Christoffel Function (CF) \cite{lasserre2019empirical}, a concept from approximation theory. By introducing CF-based loss, that is differentiable, during training, CLOE learns representations tailored for defining compact normal data supports, enabling robust outlier detection by the subsequent CF-based anomaly detection method applied to the latent space. Moreover, a particular advantage of the CF method is that it only requires one hyperparameter to be set. CLOE is computationally lightweight and designed to operate on CPUs, which is well-suited for resource-constrained environments. This method has been developed in an industry context and will be trained in a lot of different high-dimensional datasets. It requires the less computational resources possible to be trained and inferred.

The main contributions of this paper are summarized as follows:
\begin{enumerate}
  \item A new anomaly detection method, named CLOE, which performs effective representation learning in a lower-dimensional latent space guided by a training loss regularized by the empirical CF; 
  \item A differentiable implementation of the CF, using a lightweight computational approach that does not require GPU acceleration;
  \item A process for selecting the unique hyperparameter of the model, eliminating the need for extensive hyperparameter tuning;
  \item A comprehensive set of experiments on 15 high dimensional tabular datasets from the ADBench benchmark that showcase how the CLOE method outperforms several state-of-the-art baselines.
\end{enumerate}

\section{Related work} \label{sec:relatedWork}

AD methods can be classified into two different types: classical and the deep learning AD methods.

A classic way to detect outliers in a cloud of points is to estimate density, like Density-Based Spatial Clustering of Applications with Noise (DBSCAN) \cite{ester1996density}, Kernel Density Estimation (KDE) \cite{parzen1962estimation}, or Empirical Cumulative Distribution for Outlier Detection (ECOD) \cite{li2022ecod}. After density estimation, data points within low density regions are considered outliers. Another approach is to compute the distribution support used to define the boundary of normal data like One-Class Support Vector Machine (OC-SVM) \cite{scholkopf1999support}, Support Vector Data Description (SVDD) \cite{tax2004support}, and the empirical CF \cite{lasserre2019empirical}. After support computation, data points lying outside the support are then considered as outliers. A simpler method can be to compute the distance between the k-nearest neighbors (kNN) \cite{ramaswamy2000efficient} of each sample and consider those with the largest distances as outliers. However, these classical AD methods do not scale well with high-dimensional data. For example, the computation time can become prohibitively high for the empirical CF \cite{ducharlet2024leveraging}, or interdependencies between dimensions are lost in ECOD \cite{han2022adbench}.

To address these challenges, deep neural network (DNN) AD methods have been developed. Most of these approaches are semi-supervised, trained only on normal samples. The DNN is trained to reconstruct the input sample and the outlier score is computed as the difference between the input and the reconstructed output. These methods can be more complex and train a neural network to reduce data dimensionality and then feed reduced data into a classical AD method to identify the outliers. DeepSVDD \cite{ruff2018deep} extends the SVDD method by learning useful data representations and optimizing the SVDD objective. Adaptations of Deep-Clustering (DEC) \cite{xie2016unsupervised} have led to deep clustering-based anomaly detection methods: the AE is first pretrained with the reconstruction error, and then training continues with a clustering-based loss. DEC proposes a k-means-based loss \cite{xie2016unsupervised} while Deep-Clustering Compact (DCC) \cite{arellano2021deep} utilizes an OC-SVM-based loss. These methods construct new representations of the data points and then fed them into a classical AD method. However, these newly learned representations may lose relevant information for AD, making the classical AD method less effective \cite{pang2021deep}.

A solution is joint training, where the autoencoder is trained with a loss that combines the reconstruction error and a loss term from the downstream classical AD method. This approach guides representation learning and improves AD performance. The Deep Clustering Hierarchical AutoEncoder (DCVAE) and Deep Nested Clustering AutoEncoder (DNCAE) \cite{nguyen2024deep} extend deep-clustering methods by using either a double autoencoder or different layers of the same autoencoder to produce multiple representations of the data. These representations are used  to compute a k-means clustering-based loss summed with the reconstruction error. The Deep Autoencoder Gaussian Mixture Model (DAGMM) \cite{zong2018deep} combines the reconstruction error of the autoencoder with the latent space representation to feed a neural network that outputs the mixture membership predictions for each data point. The parameters of the GMM are then estimated, and the energy of each sample is computed. The model is jointly trained by optimizing the reconstruction error and the sample energy. 

The central idea of this paper is to use a CF-based method as the downstream AD method because it  offers theoretical proofs for support estimation and outlier detection. However, this method does not scale to high dimensional data. A deep neural network is used to reduce high-dimensional data, with a joint training guided by the CF, to propose data representations adjusted for the CF-based anomaly detection. 

\section{Background} \label{sec:background}
The CF is a well-known concept in approximation theory. Recent studies \cite{lasserre2019empirical}, and \cite{lasserre2022christoffel} propose to adapt it to data analysis as a means to estimate the support of a distribution, which may be highly nonlinear. This section resumes some important definitions about the CF and its empirical counterpart from~\cite{lasserre2019empirical} and~\cite{lasserre2022christoffel}.

\subsection{Presentation of the Christoffel function}
Let $\Omega \subset \mathbb{R}^d$ be a compact set with non-empty interior. Let $\mu$ be a finite Borel measure supported on $\Omega$. $\mu$ is absolutely continuous w.r.t. Lebesgue measure on $\Omega$, a set with non-empty interior and positive density. Let $v_n(x):= (P^{\alpha})_{\alpha \in \mathbb{N}^d}$ be the monomial basis of the vector space of $\mathbb{R}[x]$ of all the monomials of degree less than or equal to $n$ graded in the lexicographic order\footnote{lexicographic order: monomial are first sorted by degree and then using lexicographic order on variables considering $X_1=a$, $X_2=b$, etc.}. The size of the vector $v_n(x)$, denoted as $s_d(n)$, is equal to $\binom{d+n}{n}$.
Let $M_n(\mu)$ be the moment matrix of $\mu$. $M_n(\mu)$ is a real symmetric matrix, $M_n(\mu)$ can be written as 
\begin{equation}
     M_n(\mu) = \int_{\mathbb{R}^d}v_n(x) v_n(x)^Td\mu(x)
\end{equation}
$M_n(\mu)$ is positive definite and is non-singular for all $n$ .

For our study, we will consider the inverse of the CF. Let us introduce the Christoffel-Darboux kernel $K_n^{\mu}$ associated with $\mu$. Given any basis of $\mathbb{R}_N[x]$, orthonormal with respect to the inner product induced by $M_n(\mu)$, $(p_i)^{s_d(n)}_{i=1}$, $K_n^{\mu}$ is defined as:
\begin{equation}
    (x,y) \mapsto K_n^{\mu}(x,y):= \sum_{i=1}^{s_d(n)} p_i(x)p_i(y).
\end{equation}
This kernel can also be computed from the moment matrix:
\begin{equation}
    (x,y) \mapsto K_n^{\mu}(x,y):=v_n(x)^T M_n(\mu)^{-1}v_n(y).
\end{equation}
\begin{definition}[The Christoffel Function inverse]
The inverse of the Christoffel Function (CF) of degree $n \in \mathbb{N}$ associated with the measure $\mu$, denoted by $\Lambda_n^{\mu}(x)^{-1}$, is defined as
\begin{equation}
\Lambda_n^{\mu}(x)^{-1} := K_n^{\mu}(x,x), \forall x \in \mathbb{R}^d
\end{equation}
\end{definition}
$\Lambda_n^{\mu}(x)^{-1}$ is a sum-of-squares polynomial of degree $2n$, it is differentiable on $\mathbb{R}^d$. \cite{pauwels2016sorting} showed that $\Lambda_n^{\mu}(x)^{-1}$ has a higher value for data points that are isolated from the other points.
\subsection{The empirical Christoffel function for data analysis}
Let $\mathbb{X}\subset \mathbb{R}^D$ be a finite set of data of size $N$, $d<D$. Let $\mathbb{X}_e\subset \mathbb{R}^d$ be the encoded version of $\mathbb{X}$ in the $d$ dimension space. We consider the discrete measure $\mu_N$ whose support is $\mathbb{X}_e$ sampled from a theoretical measure $\mu$ supported on $\Omega$. The empirical version of the moment matrix can be written as 
\begin{equation}
    M_n(\mu_N) = \frac{1}{N}\sum_{z \in\mathbb{X}_e}v_n(z) v_n(z)^T.
\end{equation}

To guarantee the invertibility of the matrix $M_n(\mu_N)$, the size of $\mathbb{X}_e$ must be greater than $s_d(n)$ according to \cite{lasserre2022christoffel}, Corollary 6.3.5. Under the condition $|\mathbb{X}_e|=N >s_d(n)$, the inverse of the empirical CF is defined as:

\begin{equation}
    \Lambda_n^{\mu_N}(z)^{-1} := v_n(z)^T M_n(\mu_N)^{-1}v_n(z), z \in \mathbb{X}_e.
\end{equation}

\subsection{Thresholding with the empirical Christoffel function}
Outlier detection via the CF requires a thresholding policy. The CF is known to have theoretical properties in the analysis of discrete data to define level sets that capture quite accurately the geometric shape of the support \cite{lasserre2022christoffel}. 
Reference~\cite{lasserre2022christoffel} considers a problem in $\mathbb{R}^2$ in Chapter 7 and proposes to fix the constant $n_N$ related to this problem introduced by \cite{vu2022rate} to $n_N:=\lfloor2N^{1/4}\rfloor$. Then the empirical CF is evaluated at each point and the smallest value is chosen as threshold. This smallest value corresponds to the closest level set of the support of the normal distribution.

Reference~\cite{ducharlet2024leveraging} proposes a method, named DyCF, to detect outliers in data streams. The approach uses the Sherman-Morrison formula \cite{sherman1950adjustment} to update the moment matrix for each new data point, avoiding the need to recompute its inverse at every step. The DyCF method requires only a single hyperparameter: the polynomial degree $n$. A scoring function is then defined as:
\begin{equation}
    S_{n,d}(x)=\frac{ \Lambda_n^{\mu_N}(x)^{-1}}{\gamma_{n,d}},
\end{equation}
where $\gamma_{n,d}=Cn^{3d/2}$. A point $x$ is detected as an outlier if $S_{n,d}\geq1$. 

A second method, named DyCG, proposes a solution free of hyperparameter tuning, that leverages the growth property of the CF. In DyCG, the scoring function is derived from the DyCF computation for $n=2$ and $n=6$.

All the above thresholding scheme are performed on a low-dimensional dataset.

\section{The CLOE Method} \label{sec:method}

\begin{figure*}[ht]
\includegraphics[width=01\linewidth]{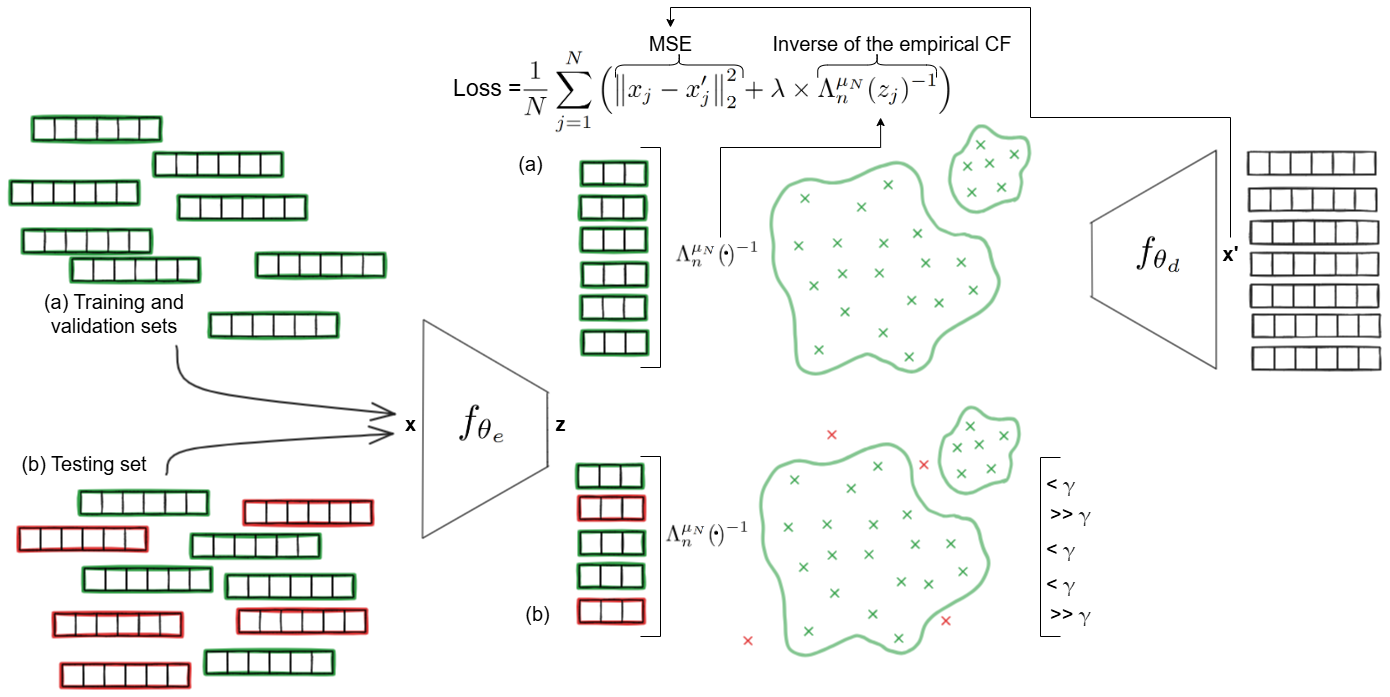} 
\caption{(a) Graphical representation of the joint training step of CLOE. The autoencoder, $f_{\theta_e}$ and $f_{\theta_d}$, is trained with normal data (green samples in the figure) to minimize the reconstruction error regularized by the inverse of the empirical Christoffel function (CF) computed on the latent space. The support of the latent space distribution is estimated with the empirical CF for all data points. (b) Graphical representation of the outlier detection step. Data points outside the support (in red) have CF values that increase exponentially with the hyperparameter $n$, much higher than the threshold $\gamma$, they are labeled as outliers.}
\label{fig:scheme}
\end{figure*}

CLOE (“Christoffel LOss for autoEncoder") is proposed as a method to utilize the inverse of the empirical CF to detect outliers in high-dimensional datasets. CLOE jointly learns a new representation of the dataset in a low-dimensional space with an AE, regularized using the empirical CF in latent space. The proposed method has four different steps. The first three steps are dedicated to the training steps, their pseudo-algorithms are detailed in Algorithm~\ref{alg:training}. The last step corresponds to the inference or anomaly detection step, its pseudo-algorithm is detailed in Algorithm~\ref{alg:inference}.

 Let $\mathbb{X}_{{train}}$ be the training set, $\mathbb{X}_{{valid}}$ be the validation set, and $\mathbb{X}_{{test}}$ be the testing set. The training and validation sets contain only normal samples (in green on Fig.~\ref{fig:scheme}).
Let $f_{\theta_e}:\mathbb{X} \subset \mathbb{R}^D\rightarrow{\mathbb{X}_e}\subset\mathbb{R}^d$ and $f_{\theta_d}:\mathbb{X}_e\rightarrow{\mathbb{R}^D}$ be the encoder and the decoder neural networks, where $\theta_e$ and $\theta_d$ are learnable parameters. 
 Let $\mathbb{X}_{e_{train}} = f_{\theta_e}(\mathbb{X}_{{train}}) $ be the encoded training set, $\mathbb{X}_{e_{valid}} = f_{\theta_e}(\mathbb{X}_{{valid}}) $ be the encoded validation set, and $\mathbb{X}_{e_{test}} = f_{\theta_e}(\mathbb{X}_{{test}}) $ be the encoded testing set.

The training is performed in three steps. The two first ones, pretraining and joint training, are for the AE training. Then the final training step corresponds to the support computation for the whole encoded training dataset and the definition of a threshold.

 \textbf{Training step 1: pretraining.} The AE is trained only for reconstruction to initialize the network weights. The loss function is the Mean Square Error (MSE):
 \begin{equation}\label{eq:reconstructionLoss}
     MSE : \frac{1}{N} \sum_{j=1}^N \left\|x_j-f_{\theta_d}\left(f_{\theta_e}\left(x_j\right)\right)\right\|^2_2 
      =\frac{1}{N} \sum_{j=1}^N \left\|x_j-x'_j\right\|^2_2.
\end{equation}

 \textbf{Training step 2: joint training.} The joint training step of the model is illustrated in Fig.~\ref{fig:scheme} (a). The training of the AE is completed with a regularized loss that combines the reconstruction loss with an empirical CF-based loss:
  \begin{align}
 \label{eq:ChristoffelLoss}
     Loss : & \frac{1}{N} \sum_{j=1}^N\left( \left\|x_j-f_{\theta_d}\left(f_{\theta_e}\left(x_j\right)\right)\right\|^2_2 
     +\lambda \times \Lambda_n^{\mu_N}(f_{\theta_e}\left(x_j\right))^{-1} \right) \\
     =& \frac{1}{N} \sum_{j=1}^N \left(\left\|x_j-x'_j\right\|^2_2
      +\lambda \times \Lambda_n^{\mu_N}(z_j)^{-1} \right),
\end{align} 
where $\lambda$ is a dynamic regularization term that controls the strength of the Christoffel loss term. $\lambda$ is computed at each epoch for each batch as the quotient of the gradient norm of the MSE loss and the gradient norm of the CF loss when this latter gradient is non-zero. This dynamic regularization ensures that the MSE loss has an impact throughout the epoch even if the Christoffel score is very high. The support of the training dataset in the latent space $\mathbb{X}_{e_{train}}$ is computed using 80\% of each batch extracted from the training dataset $\mathbb{X}_{train}$. To obtain a good estimation of the support, the number of data points to compute the support must be at least $s_d(n)$ \cite{lasserre2022christoffel}. After the support estimation, the CF is computed for all the training data and the mean of these values is utilized in the loss. This empirical CF-based loss helps the encoder to learn inlier representations more suitable for support estimation. As the value of the CF is small for samples in the support, the AE learns representations that put training data in the support of the distribution. 

 To have lower computational complexity and more stability, the Cholesky inversion method is used to invert the moment matrix. This matrix is positive definite, with the condition on $n$ and the size defined for the batch, the singularity of this matrix does not need to be checked before inversion. To avoid instability during this inversion due to large values in $\mathbb{X}_{e}$, data are normalized between $[-1,1]^d$ at the end of the encoder, i.e., $\mathbb{X}_{e} \subseteq{[-1,1]^d}$.

 \textbf{Process to choose the hyperparameter $n$.} First, $n_{max}\in \mathbb{N}$ is chosen according to $s_d(n)<|\mathbb{X}_{e_{train}}|\times 0.8$. This ensures that at least one batch is large enough to properly invert the moment matrix used to compute the support. Then, a validation step is performed at the end of each epoch. The support of the distribution in the latent space is computed with all the training data, then the CF value of each sample of the validation set is computed. The mean of all the CF values of the validation set is used to compute the loss. This loss value is monitored after the first five epochs, and the current value of $n$, starting with $n_{max}$, is validated if the loss value decreases for the following epochs during training. If the loss does not decrease, the value of $n$ is changed to $n-1$.

\textbf{Training step 3: Final support computing and threshold estimation}. The last step of the training is to encode the full training set. Then, the support of the CF is computed. A new $n_{f} \geq n$ is computed according to the condition that $s_{d}(n_{f}) <|\mathbb{X}_{e_{train}}|$. This parameter is computed according to the heuristic proposed in \cite{vu2022rate}. Then the threshold is set as:
 \begin{equation}
     \gamma_{n_{f}} = \max \{ \Lambda_{n_{f}}^{\mu_N}(z)^{-1}, z \in \mathbb{X}_{e_{train}} \}
 \end{equation}

\textbf{Inference / anomaly detection}. Fig.~\ref{fig:scheme} (b) proposes a graphical representation of this step. For a new test sample $x_{test}$, its latent representation $z_{test}=f_{\theta_e}(x_{test})$ and the Christoffel value $\Lambda_{n_{f}}^{\mu_N}(z_{test})^{-1}$ are computed. If $\Lambda_{n_{f}}^{\mu_N}(z_{test})^{-1} \leq \gamma_{n_{f}}$, then $x_{test}$ is an inlier; otherwise, $x_{test}$ is an outlier.

\begin{algorithm}
\caption{Training of CLOE}\label{alg:training}
 \DontPrintSemicolon
\SetKwInOut{Input}{Input}\SetKwInOut{Output}{Output}
\SetKw{pretraining}{Pretraining:} \SetKw{jointtraining}{Joint training:} \SetKw{final}{Support computing and threshold estimation:}
\Input{$n$, $n_{f}$, $\mathbb{X}_{train}$, $\mathbb{X}_{valid}$, $epoch_{pre}$, $epoch_{join}$}
\Output{Trained encoder $f_{\theta_e}$, trained CF $\Lambda_n^{\mu_{N}}(.)^{-1}$, threshold $\gamma_{n_{f}}$}

$d \gets$ dimension of the latent space\;
$\frac{bs \times 80}{100}  > s_d(n) $ \tcc*[r]{Batch size}
$N \gets |\mathbb{X}_{train}|$\tcc*[r]{Training set size}
\pretraining{\;
Initialize model parameters $\theta_e$ and $\theta_d$\;
    \For{each $epoch_{pre}$}{
        \For{each batch $x_{t} \subset \mathbb{X}_{train}$ }{
            $x'_{t}\gets f_{\theta_d}(f_{\theta_e}(x_{t}))$\;
            Compute MSE Loss $L_{MSE}$ \;
            Update $\theta_e$ and $\theta_d$ using gradient descent\;
        }
    }
}
\jointtraining{\;
    \For{each $epoch_{join}$}{
        \For{each batch $x_{t}\subset \mathbb{X}_{train}$}{
            $z_{t}\gets f_{\theta_e}(x_{t})$\;
            $x'_{t}\gets f_{\theta_d}(z_{t})$\;
            Randomly split $z_{t}$ in 80\% $(z_{t})^{80}$ and 20\% $(z_{t})^{20}$ sets\;
            Compute the support of the CF $\mu_{0.8bs}$ with $(z_{t})^{80}$ set\;
            Compute the CF values $(\Lambda_n^{\mu_{0.8bs}}(z_{t})^{-1})$ \;
            Compute the CLOE loss~(\ref{eq:ChristoffelLoss}) \;
            Update $\theta_e$ and $\theta_d$ using gradient descent\
        }
        $z_{i_t}\gets f_{\theta_e}(x_{i_t}), \forall i \in [1,N] $\;
        Compute the support of the CF $\mu_{N}$ with $(z_{i_t})_{1 \leq i \leq N}$\;
        \For{each validation sample $x_{v} \in \mathbb{X}_{valid}$}{
            $z_v\gets f_{\theta_e}(x_v)$\;
            $x'_v\gets f_{\theta_d}(z_v)$\;
            Compute the CF value of $z_v$, $\Lambda_n^{\mu_{N}}(z_v)^{-1}$\;
            Compute the CLOE loss~(\ref{eq:ChristoffelLoss})\;
        }
        Display the mean of all the validation loss\;
        \If{The validation does not decrease through the epochs}{
            Stop training\;
            $n \longleftarrow n-1$\;
            Start training again from pretraining step\;
        }
    }
}
\final{\;
    $z_{i_t}\gets f_{\theta_e}(x_{i_t}), \forall i \in [1,N]$\;
    Compute the support of the CF $\mu_{N}$ with $(z_{i_t})_{1 \leq i \leq N}$\;
    Compute the CF value of each training sample $(z_t)_{1 \leq i \leq N}$, $(\Lambda_{n_{f}}^{\mu_{N}}(z_{i_t})^{-1})_{1 \leq i \leq N}$\;
    $\gamma_{n_{f}}  \gets \max_{1 \leq i \leq N} (\Lambda_{n_{f}}^{\mu_{N}}(z_{i_t})^{-1})$ \;
    \Return{ $f_{\theta_e}$, $\Lambda_{n_{f}}^{\mu_{N}}(.)^{-1}$, $\gamma_{n_{f}}$}\;
}
\end{algorithm}

\begin{algorithm}
\DontPrintSemicolon
\caption{Inference and outlier detection with CLOE}\label{alg:inference}
\SetKwInOut{Input}{Input}\SetKwInOut{Output}{Output}
\Input{Trained autoencoder $f_{\theta_e}$, trained CF $\Lambda_n^{\mu_{N}}(.)^{-1}$, threshold $\gamma_{n_{f}}$, test sample $x_{test}$}
\Output{ 0 (inlier) or 1 (outlier) }
$z_{test} \gets f_{\theta_e}(x_{test})$\;
Compute the CF value of the encoded test sample: $\Lambda_{n_{f}}^{\mu_{N}}(z_{test})^{-1}$\;
\eIf{$\Lambda_{n_{f}}^{\mu_{N}}(z_{test})^{-1} \leq \gamma_{n_{f}}$}
{\Return{0} \tcc*{$x_{test}$ in an inlier}}
{\Return{1}  \tcc*{$x_{test}$ is an outlier}}

\end{algorithm}

\section{Experiments} \label{sec:experiments}

\subsection{Datasets}
To evaluate the CLOE method, we use several datasets from ADBench \cite{han2022adbench}. This benchmark provides a diverse collection of datasets for anomaly detection with distinctive features. As our focus is on high dimensional data and not only images, we selected 15 datasets, each with dimensions between 9 and 1555. The number of data points per dataset varies between 80 and 299285.

For each dataset, outliers are utilized exclusively during the testing step. The inlier dataset is divided into a training (70\%), validation (20\%) and testing (10\%) set. To compare our results against different baseline methods, we fix a random seed to produce identical splits across experiments.

\subsection{Baseline methods}

Our method is compared to DAGMM~\cite{zong2018deep},  OC-SVM~\cite{scholkopf1999support}, iForest~\cite{liu2008isolation}, ECOD~\cite{li2022ecod}, Deep SVDD~\cite{ruff2018deep}, kNN~\cite{ramaswamy2000efficient} and KDE~\cite{parzen1962estimation}. Only methods that can be trained and tested without GPU acceleration are considered\footnote{Experiments with other different methods have been conducted and are available in the supplemental work.}. For DAGMM, we use the implementation proposed in \cite{han2022adbench}. Then, for the last six models, we use the PyOD implementations \cite{zhao2019pyod}. The hyperparameters of all baselines are set according to the corresponding original papers.

DAGMM \cite{zong2018deep} is the method most similar to CLOE. However, DAGMM uses a neural network to predict the sample mixture membership. The model is an adaptation of the mixture model. It differs from CLOE, which directly applies AD methods in its loss instead of adapting them. 

DeepSVDD \cite{ruff2018deep} is also a method with a deep neural network, similar to CLOE. However, the AE and the AD model are trained separately. The representations of the data may not be well-suited for SVDD.

OC-SVM \cite{scholkopf1999support}, iForest \cite{liu2008isolation}, ECOD \cite{li2022ecod}, kNN \cite{ramaswamy2000efficient} and KDE \cite{parzen1962estimation} are classical AD methods that do not rely on deep neural networks. They are computationally efficient, but may struggle to achieve high performance on high-dimensional datasets.

\subsection{Evaluation metrics}

We evaluate our results using Area Under the Receiver Operating Characteristic curve (AU-ROC) and Average Precision Area Under Curve (AP AUC), the same metrics used in the ADBench paper \cite{han2022adbench} to compare the different methods. Both metrics are computed using the implementation provided by the scikit-learn Python package~\cite{scikit-learn}. The AU-ROC metric reflects the trade-off between true positive and false positive rates. AP AUC combines precision and recall metrics. It is particularly informative for imbalanced data, which is the case with all the datasets, as there are few outliers. 

\subsection{Implementation}
 As in \cite{xie2016unsupervised}, the AE has 3 hidden layers of dimensions 500, 500, and 2000, using ReLU activation functions. The latent space dimension is set to $d=8$, chosen according to the complexity of computing the moment matrix of the training set with $2\le n\le 6$. A dropout rate of 20\% is applied for the pretraining step and no dropout is used for the joint training step according to the configuration proposed in \cite{xie2016unsupervised}. At the end of the encoder, a batch norm layer followed by a Hyperbolic Tangent (Tanh) activation layer is added to ensure the encoded data lie within $[-1;1]$. This condition is required to compute the moment matrix and invert it using the Cholesky algorithm\footnote{\label{fn:code}Code and supplementary work available at: \url{https://gitlab.laas.fr/lbillet1/cloe}.}.

The pretraining phase is conducted for 10 epochs with an early stopping rule based on the value of the validation loss. The joint training is conducted for 150 epochs with an early-stopping policy of 10 epochs. The Adam optimizer is used with a learning rate of $10^{-4}$, $\beta_{1}=0.9$, $\beta_{2}=0.999$, and the Glorot initialization for all datasets. All experiments were conducted on a device with 8 CPUs and 32 GB RAM.  The training time varies between 33 and 3600 seconds depending of the sizes of the training dataset and the batch. The inference time for one sample is on the order of $10^{-3}$ seconds.

The pretraining and joint training steps are conducted in batches, with batch size $bs \times 0.8  > s_d(n)$.

The value of $n$ for each dataset, a sensitivity analysis of the latent space dimension, of the number of epochs, and of the value of $n$ for two datasets, and the training and inference time for each dataset of CLOE are in the supplementary work\textsuperscript{\ref{fn:code}}.

\subsection{Results}
\begin{table*}
\caption{AU-ROC for the different methods on the selected datasets, with variances}
\label{t:AUROCVar} \fontsize{7}{7}\selectfont
\begin{tabular}{c|cccccccc}
\hline
Dataset & CLOE &DAGMM& OC-SVM & iForest & ECOD& Deep SVDD & kNN&  KDE\\ 
\hline
 \textit{ALOI} &\textbf{0.561} $(\pm$3e-5)&0.529 $(\pm $1e-4)&0.517 $(\pm$2e-8)&0.539 $(\pm$6e-6)&0.531 $(\pm$1e-9)&0.546 $(\pm$1e-4)&\underline{0.556} $(\pm$4e-6)&0.518 $(\pm$1e-8)\\
 \textit{backdoor} &\textbf{0.944} $(\pm$1e-2)&0.619$(\pm $3e-3)&0.865 $(\pm $3e-6) & 0.750 $(\pm $9e-4) &0.846 $(\pm $7e-9) &0.553 $(\pm $1e-3)&\underline{0.938} $(\pm $7e-7)&0.915 $(\pm $2e-6)\\   
 \textit{breastw} &0.994$(\pm $5e-6)& N/A & \underline{0.997} $(\pm $2e-7)& 0.994 $(\pm $1e-7)&0.994 $(\pm $9e-10)& 0.988 $(\pm $2e-5)&0.995 $(\pm $3e-7)& \textbf{0.998} $(\pm $3e-7)\\   
 \textit{campaign} &0.610 $(\pm $3e-4)&0.603 $(\pm $6e-4)&0.689 $(\pm $1e-5)&0.721 $(\pm $1e-4)&\textbf{0.772} $(\pm $4e-8)&0.710 $(\pm $1e-3)&\underline{0.725} $(\pm $3e-6)& 0.699 $(\pm $2e-6)\\
 \textit{cardio} &\textbf{0.979} $(\pm $1e-3)&0.527 $(\pm $5e-4)&0.957 $(\pm $2e-7)&0.951 $(\pm $6e-5)&0.946 $(\pm $3e-7)&0.953 $(\pm $2e-3)&0.933 $(\pm $8e-6)&\underline{0.977} $(\pm $3e-6)\\ 
 \textit{census} &\textbf{0.715} $(\pm $1e-4)&0.605 $(\pm $2e-4)&0.553 $(\pm $2e-5)&0.611 $(\pm $9e-4)&0.659 $(\pm $4e-10)&\underline{0.702} $(\pm $2e-4)&0.661 $(\pm $8e-6)&0.662 $(\pm $3e-6)\\
 \textit{fault} &\textbf{0.928} $(\pm $9e-6)&0.496 $(\pm $2e-3)&0.591 $(\pm $3e-6)&0.662 $(\pm $9e-5)&0.485 $(\pm $7e-7)&0.542 $(\pm $3e-3)&0.822 $(\pm $4e-6)&\underline{0.884} $(\pm $4e-6)\\ 
 \textit{Hepatitis} &\textbf{0.938} $(\pm $2e-4)&0.589 $(\pm $6e-3)&\underline{0.855} $(\pm $7e-5)&0.816 $(\pm $1e-4)&0.786 $(\pm $3e-5)&0.789 $(\pm $1e-3)& 0.639 $(\pm $7e-4)&\underline{0.855} $(\pm $8e-5)\\
 \textit{InternetAds} &\textbf{0.878} $(\pm $2e-4)&N/A&0.708 $(\pm $6e-7)&0.425 $(\pm $7e-4)&0.698 $(\pm $1e-7)&0.749 $(\pm $1e-3)&\underline{0.823} $(\pm $2e-5)&0.815 $(\pm $1e-6)\\  
 \textit{landsat} &\textbf{0.854} $(\pm $8e-4)&0.580 $(\pm $8e-3)&0.471 $(\pm $9e-7)&0.614 $(\pm $7e-4)&0.388 $(\pm $3e-7)&0.462 $(\pm $1e-3)&\underline{0.784} $(\pm $3e-6)&0.757 $(\pm $2e-6)\\ 
 \textit{letter} &0.943 $(\pm $7e-5)&0.391 $(\pm $6e-4)&\underline{0.977} $(\pm $5e-6)&0.639 $(\pm $7e-5)&0.579 $(\pm $4e-7)&0.523 $(\pm $2e-3)&0.917 $(\pm $9e-6)&\textbf{0.980} $(\pm $7e-6)\\ 
 \textit{mnist} &0.871 $(\pm $5e-5)&0.615 $(\pm $4e-4)&0.789 $(\pm $0)&0.860 $(\pm $4e-4)&0.768 $(\pm $6e-7)&0.834 $(\pm $2e-3)&\textbf{0.937} $(\pm $9e-6)&\underline{0.920} $(\pm $2e-5)\\
 \textit{musk} &\textbf{1.0} $(\pm $0)&0.485 $(\pm $2e-2)&0.859 $(\pm $0)&0.960 $(\pm $6e-4)&0.993 $(\pm $3e-8)&0.998 $(\pm $5e-6)&\textbf{1.0} $(\pm $0)&\textbf{1.0} $(\pm $0)\\
 \textit{shuttle} &\textbf{0.998} $(\pm $5e-3)&0.991 $(\pm $2e-3)&\underline{0.997} $(\pm $6e-9)&0.996 $(\pm $8e-8)&0.993 $(\pm $1e-9)&0.994 $(\pm $6e-6)&0.995 $(\pm $6e-9)&\underline{0.997} $(\pm $5e-8)\\ 
 \textit{speech} &\underline{0.859} $(\pm $1e-4)&0.489 $(\pm $2e-4)&0.469 $(\pm $5e-7)&0.479 $(\pm $2e-4)&0.473 $(\pm $2e-9)&0.508 $(\pm $3e-4)&0.501 $(\pm $1e-5)&\textbf{0.881} $(\pm $1e-4)\\  
\hline
Mean &\textbf{0.871}&0.578&0.753&0.734&0.727&0.723&0.815&\underline{0.857}\\
 Rank &\textbf{1}&8&4&5&6&7&3&\underline{2}\\
 \hline
\end{tabular}
\end{table*}
\begin{table*}
\caption{AP AUC for the different methods on the selected datasets, with variances}
\label{t:APAUCVar} \fontsize{7}{7}\selectfont
\begin{tabular}{c|cccccccc}
\hline
Dataset & CLOE &DAGMM& OC-SVM & iForest & ECOD& Deep SVDD & kNN&  KDE\\ 
\hline
 \textit{ALOI} &\underline{0.044} $(\pm $9e-6)&0.041 $(\pm $2e-5)&0.041 $(\pm $2e-9)&0.033 $(\pm $9e-8)&0.032 $(\pm $4e-11)&0.037 $(\pm $1e-6)&\textbf{0.049} $(\pm $7e-6)&0.042 $(\pm $5e-10)\\  
 \textit{backdoor} &\textbf{0.745} $(\pm $2e-2)&0.033 $(\pm $5e-4)&0.107 $(\pm $1e-5)&0.048 $(\pm $6e-5)&0.093 $(\pm $1e-8)&0.038 $(\pm $1e-4)&\underline{0.517} $(\pm $1e-3)&0.411 $(\pm $4e-6)\\   
 \textit{breastw} &0.985 $(\pm $6e-5)&N/A&\underline{0.994} $(\pm $1e-6)&0.989 $(\pm $6e-7)&0.987 $(\pm $5e-9)&0.973 $(\pm $1e-4)&0.991 $(\pm $1e-6)&\textbf{0.996} $(\pm $2e-6)\\  
 \textit{campaign} &0.178 $(\pm $3e-6)&0.177 $(\pm $2e-4)&\underline{0.310} $(\pm $9e-6)&0.302 $(\pm $1e-4)&\textbf{0.356} $(\pm $5e-8)&0.290 $(\pm $6e-4)&0.304 $(\pm $9e-6)&0.296 $(\pm $5e-6)\\
 \textit{cardio} &\underline{0.817}$(\pm $3e-2)&0.116 $(\pm $4e-3)&0.665 $(\pm $3e-5)&0.679 $(\pm $2e-3)&0.626 $(\pm $2e-5)&0.705 $(\pm $4e-3)&0.667 $(\pm $8e-5)&\textbf{0.861} $(\pm $6e-4)\\   
 \textit{census} &\underline{0.112} $(\pm $2e-3)&0.086 $(\pm $1e-4)&0.065 $(\pm $3e-7)&0.074 $(\pm $3e-5)&0.084 $(\pm $3e-11)&\textbf{0.126} $(\pm $5e-4)&0.084 $(\pm $5e-7)&0.084 $(\pm $2e-7)\\
 \textit{fault} &\textbf{0.828} $(\pm $8e-4)&0.365 $(\pm $1e-3)&0.458 $(\pm $2e-6)&0.495 $(\pm $8e-5)&0.337 $(\pm $3e-7)&0.419 $(\pm $2e-3)&0.668 $(\pm $9e-5)&\underline{0.825} $(\pm $5e-5)\\ 
 \textit{Hepatitis} &\textbf{0.670} $(\pm $3e-3) &0.214 $(\pm $6e-3)&0.395 $(\pm $3e-4)&0.400 $(\pm $2e-4)& 0.356 $(\pm $9e-5)&\underline{0.439} $(\pm $6e-3)& 0.251 $(\pm $4e-4)&0.424 $(\pm $2e-3)\\   
 \textit{InternetAds} &0.526 $(\pm $6e-3)&N/A&\underline{0.578} $(\pm $2e-6)&0.155 $(\pm $1e-4)&0.552 $(\pm $1e-6)&0.495 $(\pm $4e-3)&0.692 $(\pm $1e-5)&\textbf{0.747} $(\pm $1e-5)\\  
 \textit{landsat} &\textbf{0.739} $(\pm $4e-4)&0.267 $(\pm $5e-4)&0.199 $(\pm $2e-7)&0.273 $(\pm $5e-4)&0.172 $(\pm $2e-8)&0.195 $(\pm $2e-4)&0.473 $(\pm $3e-5)&\underline{0.499} $(\pm $3e-5)\\ 
 \textit{letter} &0.644 $(\pm $2e-3)&0.067 $(\pm $1e-5)&\textbf{0.731} $(\pm $3e-3)&0.091 $(\pm $5e-6)&0.079 $(\pm $6e-8)&0.074 $(\pm $5e-5)&0.411 $(\pm $2e-4)&\underline{0.723} $(\pm $2e-3)\\ 
 \textit{mnist} &0.519 $(\pm $1e-3)&0.170 $(\pm $6e-4)&0.194 $(\pm $0)&0.377 $(\pm $2e-3)&0.194 $(\pm $3e-7)&0.455 $(\pm $5e-3)&\textbf{0.666} $(\pm $5e-5)&\underline{0.640} $(\pm $5e-4)\\
 \textit{musk} &\textbf{0.999} $(\pm $0)&0.048 $(\pm $1e-3)&0.104 $(\pm $0)&0.472 $(\pm $5e-2)&0.855 $(\pm $1e-5)&0.941 $(\pm $3e-3)&\textbf{0.999} $(\pm $0)&\textbf{0.999} $(\pm $0)\\
 \textit{shuttle} &\textbf{0.978} $(\pm $2e-2)&0.853 $(\pm $2e-3)&0.939 $(\pm $3e-6)&\underline{0.976} $(\pm $1e-5)&0.912 $(\pm $2e-7)&0.914 $(\pm $2e-4)&0.854 $(\pm $7e-6)&0.875 $(\pm $6e-5)\\     
 \textit{speech} &\underline{0.068} $(\pm $3e-4)& 0.016 $(\pm $3e-4)&0.019 $(\pm $2e-7)&0.079 $(\pm $3e-4)&0.020 $(\pm $4e-10)&0.017 $(\pm $3e-7)&0.020 $(\pm $2e-8)&\textbf{0.118} $(\pm $7e-4)\\
  \hline
 Mean &\textbf{0.590}&0.189&0.387&0.363&0.377&0.408&0.510&\underline{0.569}\\
 Rank &\textbf{1}&8&5&7&6&4&3&\underline{2}\\
 \hline
\end{tabular}
\end{table*}

 Tables~\ref{t:AUROCVar} and \ref{t:APAUCVar} show the results for the 15 selected datasets with the metric AU-ROC and AP AUC for CLOE and its baselines. The experiments were repeated 5 times with different random seeds. The mean and variance results are presented. The highest values are in bold and the second are underlined.
 
Regarding the deep learning methods, DAGMM requires a matrix that is not always invertible, preventing successful training on some datasets (marked as N/A in the tables as the model could not be trained on the corresponding dataset). Across the test datasets, CLOE outperforms DeepSVDD, and DAGMM. 

Regarding classical AD methods, CLOE outperforms in 9 datasets according to AU-ROC and in 6 according to AP AUC. CLOE is on average better than the other classical methods on all datasets for both metrics. However, KDE obtains good performances on some datasets. As was shown in~\cite{ducharlet2024leveraging}, the visual analysis of the level sets produced by the CF-based AD method and KDE shows that the level sets of the CF-based AD method are better fitted to data distribution.

\subsection{Ablation studies}
\label{append:ablationStudy} 
\begin{table}
\caption{AU-ROC for the ablation study}
\label{t:ablationStudyROC} \fontsize{7}{6.5}\selectfont
\begin{tabular}{c|ccccc}
\hline
{\small Dataset}&{\small CLOE} &\makecell{\small No \\\small pretraining}  &  \makecell{\small No joint \\\small training} & \makecell{\small Untrained \\ \small AE} & {\small KDE}
 \\ \hline
 \textit{ALOI}&0.561&\textbf{0.577}&0.554&0.549&0.504\\
 \textit{backdoor}&\textbf{0.944}&0.922&0.851&0.826&0.540\\
 \textit{breastw}&\textbf{0.994}&\textbf{0.994}&0.929&0.986&0.992\\ 
 \textit{campaign}&\textbf{0.610}&0.439&0.561&0.604&0.591\\
 \textit{cardio}&\textbf{0.979}&0.904&0.785&0.874&0.746\\
 \textit{census}&\textbf{0.715}&0.544&0.624&0.611&0.500\\
 \textit{fault}&\textbf{0.928}&0.875&0.816&0.616&0.500\\
 \textit{Hepatitis}&\textbf{0.938}&0.901&0.618&0.802&0.500\\
 \textit{InternetAds}&\textbf{0.878}&0.862&0.725&0.625&0.871\\
 \textit{landsat}&\textbf{0.854}&0.812&0.780&0.608&0.500\\
 \textit{letter}&\textbf{0.943}&0.936&0.738&0.557&0.725\\ 
 \textit{mnist}&\textbf{0.871}&0.623&0.637&0.731&0.500\\
 \textit{musk}&\textbf{1.0}&\textbf{1.0}&0.856&0.929&\textbf{1.0}\\
 \textit{shuttle}&\textbf{0.998}&0.994&0.883&0.995&0.991\\
 \textit{speech}&\textbf{0.859}&0.650&0.463&0.508&0.679\\
   \hline
 Mean&\textbf{0.871}&0.802&0.722&0.725&0.676\\
\hline
\end{tabular}
\end{table}
\begin{table}
\caption{AP AUC for the ablation study}
\label{t:ablationStudyAP} \fontsize{7}{6.5}\selectfont
\begin{tabular}{c|ccccc}
\hline
{\small Dataset}&{\small CLOE} &\makecell{\small No \\\small pretraining}  &  \makecell{\small No joint \\\small training} & \makecell{\small Untrained \\ \small AE} & {\small KDE}
 \\\hline
 \textit{ALOI}&\textbf{0.044}&0.040&0.039&0.038&0.033\\
 \textit{backdoor}&\textbf{0.745}&0.544&0.202&0.454&0.126\\
 \textit{breastw}&0.985&\textbf{0.988}&0.854&0.975&0.978\\ 
 \textit{campaign}&\textbf{0.178}&0.103&0.141&0.169&0.161\\
 \textit{cardio}&\textbf{0.817}&0.565&0.398&0.584&0.292\\
 \textit{census}&\textbf{0.112}&0.065&0.088&0.077&0.062\\
 \textit{fault}&\textbf{0.828}&0.792&0.728&0.478&0.351\\
 \textit{Hepatitis}&\textbf{0.670}&0.550&0.363&0.477&0.158\\
 \textit{InternetAds}&0.526&0.522&0.386&0.330&\textbf{0.737}\\
 \textit{landsat}&\textbf{0.739}&0.618&0.572&0.315&0.226\\
 \textit{letter}&\textbf{0.644}&0.512&0.234&0.114&0.425\\ 
 \textit{mnist}&\textbf{0.519}&0.261&0.192&0.329&0.092\\
 \textit{musk}&\textbf{0.999}&0.998&0.522&0.616&\textbf{0.999}\\
 \textit{shuttle}&\textbf{0.978}&0.956&0.794&0.938&0.947\\
 \textit{speech}&\textbf{0.068}&0.043&0.017&0.024&0.032\\
  \hline
 Mean&\textbf{0.590}&0.504&0.369&0.394&0.374\\
 \hline
\end{tabular}
\end{table}

\begin{table*}
\caption{F1-score for different threshold}
\label{t:f1Threshold}   \fontsize{7.8}{6.5}\selectfont
\begin{tabular}{c|cccccccccc}
\hline
Dataset &Optimized &Adjusted & CLOE & 90th p train & 75th p train &50th p train&100th p valid&90th p valid & 75th p valid &50th p valid
 \\ \hline
 \textit{ALOI}&0.075&0.066&\textbf{0.075}&0.072&0.065&0.064&0.010&0.073&0.067&0.064\\
 \textit{backdoor}&0.810&0.741&\textbf{0.805}&0.269&0.131&0.077&0.627&0.227&0.122&0.072\\
 \textit{breastw}&0.949&0.941&0.937&0.880&0.793&0.670&0.222&\textbf{0.950}&0.871&0.710\\ 
 \textit{campaign}&0.248&0.194&0.231&0.245&0.236&0.223&0.008&0.186&0.233&\textbf{0.246}\\
 \textit{cardio}&0.681&0.681&0.432&0.377&0.315&0.250&0.390&\textbf{0.690}&0.636&0.553\\
 \textit{census}&0.170&0.098&0.119&0.143&0.157&\textbf{0.167}&0.007&0.125&0.155&\textbf{0.167}\\
 \textit{fault}&0.811&0.765&\textbf{0.793}&0.752&0.699&0.625&0.017&0.454&0.610&0.779\\
 \textit{Hepatitis}&0.720&0.692&0.565&0.520&0.448&0.388&0.133&0.571&0.440&\textbf{0.667}\\
 \textit{InternetAds}&0.632&0.522&\textbf{0.621}&0.562&0.500&0.418&0.016&0.393&0.496&0.549\\
 \textit{landsat}&0.741&0.708&\textbf{0.732}&0.652&0.562&0.461&0.178&0.585&0.688&0.728\\
 \textit{letter}&0.405&0.380&0.323&0.278&0.227&0.174&0.019&0.352&0.362&\textbf{0.382}\\ 
 \textit{mnist}&0.375&0.369&0.063&\textbf{0.364}&0.307&0.239&0.133&0.364&0.315&0.237\\
 \textit{musk}&1.0&1.0&0.951&0.381&0.206&0.115&\textbf{0.979}&0.421&0.207&0.114\\
 \textit{shuttle}&0.980&0.976&\textbf{0.938}&0.501&0.338&0.222&0.912&0.576&0.368&0.235\\
 \textit{speech}&0.107&0.032&\textbf{0.107}&0.087&0.068&0.049&0.0&0.044&0.068&0.096\\
  \hline
 Mean&0.580&0.544&\textbf{0.513}&0.407&0.337&0.276&0.243&0.400&0.376&0.373\\
 \hline
\end{tabular}
\end{table*}

To check the utility of each training step of the AE of our method, we performed an ablation study using all datasets.

First, we remove the pretraining step. The weights of the AE are randomly initialized and the joint training is performed until the validation loss stops improving. The joint training and the support computation steps remain unchanged from the full method. 

Second, we remove the joint training step. The AE is first trained for 10 epochs using only the reconstruction loss~(\ref{eq:reconstructionLoss}). The CF support is then computed and the threshold defined in the original method is used.

Third, \cite{ryu2024can} raises a warning concerning good performance of untrained neural network. To confirm the utility to train the AE in CLOE, an experiment with randomly initialized weights and data encoded from the latent space of the untrained AE is conducted. The CF is trained with these encoded data.

Then, CF is replaced by a kernel method, KDE, in the joint training step and in the final anomaly detection step. For the implementation, the differentiable KDE \cite{diffKDE} is used. The AE structure and all training parameters are kept from CLOE. Reference~\cite{ducharlet2024leveraging} compares CF and KDE and shows that CF obtains better performance. This experiment extends this previous test and compares KDE and CF in the latent space of an AE with guided losses.

The results are presented in Tables~\ref{t:ablationStudyROC} for the AU-ROC metric and \ref{t:ablationStudyAP} for AP AUC metric. The study shows that pretraining step and joint training are needed, as models without pretraining or without joint training step underperform compared to CLOE. For datasets of dimension 9, the performance without the joint training step is very close to CLOE performance. These results confirm that CLOE is designed for high-dimensional tabular datasets. CLOE is recommended for dimensions higher than 10. For lower dimensions, the recommendation is for CF-based AD method without AE~\cite{ducharlet2024leveraging}. The untrained AE can obtain good performance on some datasets like \textit{shuttle} or \textit{campaign}. Changing CF to KDE decreases the performance of the model. The choice of using CF instead of a kernel AD method is empirically justified. A complete training strategy improves the mean AU-ROC and AP AUC scores by approximately 7\% to 16\% and 12\% to 36\%, respectively. This highlights the importance of implementing the complete training approach.

Finally, to confirm our thresholding scheme, an experiment has been conducted with different methods to determine the threshold, using the F1-score to compare the results (cf. Table~\ref{t:f1Threshold}). The results of the "Optimized" column are obtained with a threshold iteratively optimized on the F1-score. Those of the "Adjusted" column are obtained with a threshold adjusted on the outlier contamination ratio of the test dataset. The "CLOE" column reports the results with a threshold indicated by our method. Finally, the other columns report the results for the thresholds set by quartiles of the training or validation sets, specifically 50th (median), 75th, 90th, and 100th percentiles.

On average across all datasets, the CLOE threshold achieves the best performance after the Adjusted threshold. This study shows the robustness of CLOE to determine automatically the threshold.

\section{Conclusion and future works} \label{sec:conclusion} 
In this work, we propose CLOE, an empirical CF guided AE method with a differentiable implementation of the CF, to detect outliers in high-dimensional data. Importantly, CLOE requires the tuning of only one single hyperparameter. As this hyperparameter can take only a few different values, the tuning is fast. One limitation of CLOE is that it requires a reduced dimension of the latent space, set to 8 in this work, due to the increasing size of the moment matrix to invert. The experiments show that CLOE achieves really good performance for most of the datasets. In addition, CLOE comes with an automatic threshold scheme that provides a robust way to detect outliers. Interestingly, CLOE is designed to be trained without a GPU.

Future works could extend CLOE to AD in images and sequences by modifying the autoencoder structure (e.g., using convolutional or recurrent layers). Also, analyzing the latent space with metrics like clustering quality, separability, or compactness could reveal the benefits of the CF-based loss term. Finally, the Univariate CF (UCF) was introduced as another scaling solution \cite{grivet2026scalable}, and comparing it with CLOE would be valuable.

\section*{Acknowledgment}  
The authors would like to thank Christophe Merle, Head of the Manufacturing Intelligent 5 team at Schaeffler and ANITI Industrial Coordinator, for sharing his expertise in robotized production lines. Our work has benefited from the AI Interdisciplinary Institute ANITI funded by the France 2030 program under the Grant agreements n°ANR-19-P3IA-0004 and n°ANR-23-IACL-0002.

\IEEEtriggeratref{14}
\bibliographystyle{IEEEtran}
\bibliography{IEEEabrv,ref.bib}

\end{document}